\documentclass[runningheads]{llncs}
\usepackage{graphicx}
\usepackage{marvosym}
\usepackage{indentfirst}
\usepackage{multirow}
\usepackage{rotating}
\usepackage{booktabs}
\usepackage{graphicx}
\usepackage{float}
\usepackage{subfigure}
\usepackage{amssymb}
\usepackage{amsmath}
\usepackage{xcolor}
\usepackage{enumitem}
\usepackage{makecell}
\usepackage{threeparttable}

\usepackage{hyperref}
\hypersetup{
  colorlinks,
  citecolor=blue,
  linkcolor=red,
  urlcolor=blue
}

\begin{document}
\title{Scribble2D5: Weakly-Supervised Volumetric Image Segmentation via Scribble Annotations}
\titlerunning{Scribble2D5}
%

\author{Qiuhui Chen\inst{1} \and 
Yi Hong\inst{1}\textsuperscript{(\Letter)}
}
%
\authorrunning{Chen and Hong}
%
\institute{
$^1$Department of Computer Science and Engineering, Shanghai Jiao Tong University 
\email{yi.hong@sjtu.edu.cn}\\
}
%
\maketitle 
\begin{abstract}
Image segmentation using weak annotations like scribbles has gained great attention, since such annotations are easier to obtain compared to time-consuming and labor-intensive labeling at the pixel/voxel level. However, scribbles lack structure information of the region of interest (ROI), thus existing scribble-based methods suffer from poor boundary localization. Moreover, current methods are mostly designed for 2D image segmentation, which do not fully leverage volumetric information. In this paper, we propose a scribble-based volumetric image segmentation, Scribble2D5, which tackles 3D anisotropic image segmentation and improves boundary predictions. To achieve this, we augment a 2.5D attention UNet with a proposed label propagation module to extend semantic information from scribbles and a combination of static and active boundary prediction to learn ROI's boundaries and regularize its shape. Extensive experiments on three public datasets demonstrate Scribble2D5 significantly outperforms existing scribble-based methods and approaches the performance of fully-supervised ones. Our code is available at \href{https://github.com/Qybc/Scribble2D5}{https://github.com/Qybc/Scribble2D5}. 

\end{abstract}

\section{Introduction}

Deep learning based methods have achieved impressive accuracy in many medical segmentation tasks, especially in a fully-supervised manner~\cite{shapey2019artificial,zhou2018unet++}. However, such segmentation methods typically require a large amount of dense annotations for pixels or voxels to train a deep model. While dense annotations sometimes are not easy to obtain in practice because annotating at the image pixel-/voxel-level is time-consuming and needs medical expertise to provide high-quality labels. On the other hand, fully-unsupervised segmentation methods~\cite{xia2017w,dey2021asc} have shown promising results; however, their performance gap with respect to fully-supervised approaches is too large to make them practical. Therefore, weakly-supervised approaches by using weak annotations have gained great attention to greatly reduce the workload of manual annotations while having promising and comparable results compared to fully-supervised approaches.

Commonly-used weak annotations include image-level annotations~\cite{xu2015learning,ahn2018learning}, bounding boxes~\cite{rajchl2016deepcut,shapey2019artificial}, scribbles~\cite{lin2016scribblesup,tang2018normalized,dorent2020scribble,valvano2021learning}, and extreme points~\cite{maninis2018deep,roth2021going}, etc. Compared to image-level and bounding box annotations, scribbles provide rough positions of Region of Interests (ROIs) to allow a better location of ROI. Also, scribbles are more flexible than bounding boxes and extreme points when annotating, especially for ROIs with irregular shapes. In addition, extreme point annotations are more suitable for convex shapes and may not work well for non-convex ones. Therefore, we choose scribbles as our weak annotations. 
However, scribbles are often sparse with no structure information of ROIs; as a result, scribble-based methods have difficulty in accurately locating ROI boundaries~\cite{tang2018normalized}.  
Moreover, existing methods~\cite{lin2016scribblesup,wang2019boundary,zhang2020weakly,valvano2021learning,luo2022scribble,zhang2022cyclemix} are typically designed for segmenting 2D image slices, which do not fully leverage the whole image volume, with missing continuity between slices. Researchers attempt to alleviate such issue by regularizing the volume size of segmentation outputs~\cite{kervadec2019constrained}. Another solution~\cite{dorent2020scribble} performs 3D segmentation with transfer learning, which learns with dense annotations in the source domain and with scribbles in the target domain. 


\begin{figure}[t]
\includegraphics[width=\textwidth]{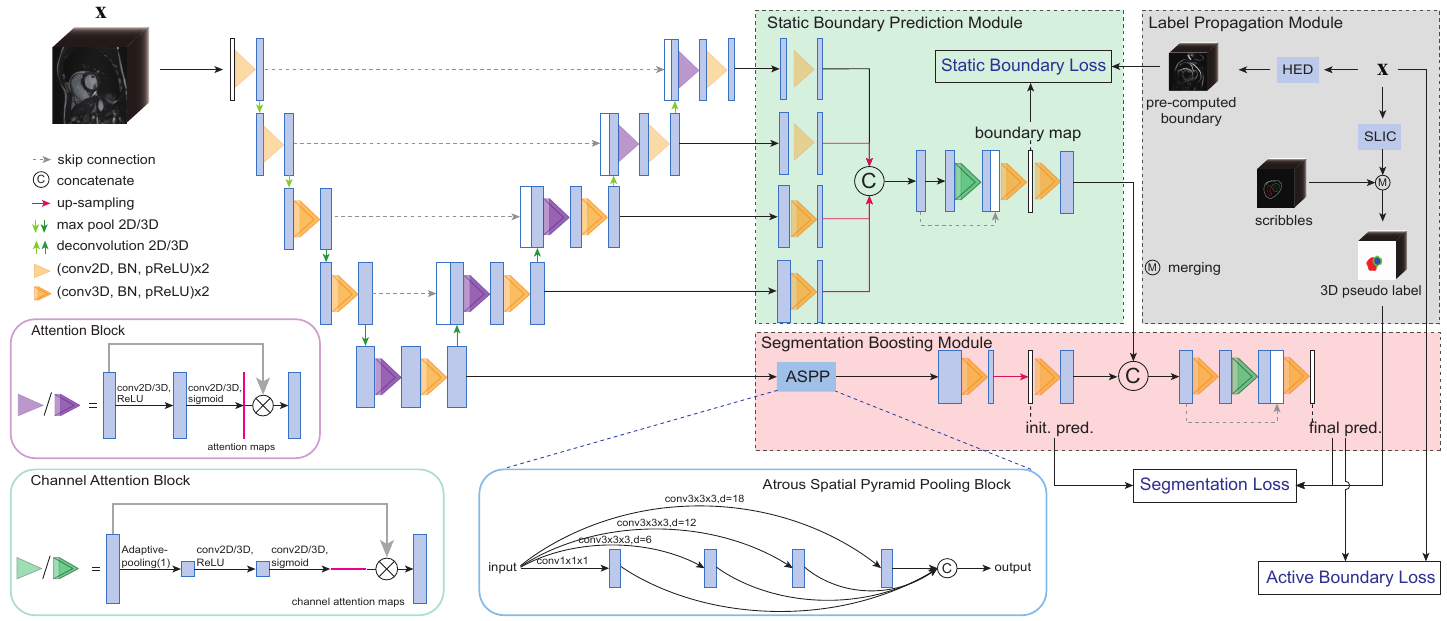}
\caption{Our Scribble2D5 architecture for volumetric image segmentation: 1) pseudo label propagation (gray box): generating pseudo 3D segmentation masks and pre-computed boundaries; 2) static boundary prediction (green box): incorporating object boundary information from the input image; 3) segmentation boosting (pink box): further considering active boundaries via an active boundary loss. (Best view in color)} 
\label{fig:overview}
\end{figure}

Inspired by two recent works~\cite{wang2019boundary,zhang2020weakly}, we propose a volumetric segmentation network based on scribble annotations, called scribble2D5. As shown in Fig.~\ref{fig:overview}, we adopt a 2.5D attention UNet~\cite{shapey2019artificial} to handle anisotropic medical volumes with different voxel spacings, which is very common in practice like our datasets. To amplify the influence of sparse scribbles and suit for volumetric segmentation, we use a label propagation module based on supervoxels to generate 3D psuedo masks from scribbles for supervision. To address the boundary localization issue, we propose to learn both static and active boundaries via predicting edges in 3D and optimizing an active boundary loss in 3D based on active contour model~\cite{chen2019learning}. 
Different from existing solutions, our scribble2D5 tackles 3D anisotropic image inputs directly, and needs scribbles only for training. At the inference stage, segmentation operates automatically, with no need of scribble inputs.

We evaluate our methods on three datasets, including ACDC dataset~\cite{bernard2018deep} for cardiac segmentation, VS dataset~\cite{shapey2021segmentation} for tumor segmentation, and CHAOS dataset~\cite{kavur2021chaos} for abdominal organ segmentation. For both ACDC and CHAOS datasets, our method outperforms the current state-of-the-art (SOTA) by large margins on three different evaluation metrics; and on the VS dataset, our method achieves better performance in Dice compared to SOTA. Our method reduces the performance gap between weakly-supervised and fully-supervised approaches, which makes it more practical to be used in the future. 

\vspace{0.05in}
\noindent
Overall, our contributions in the paper are summarized as follows:
\begin{itemize}[noitemsep,nolistsep]
    \item We propose a scribble2D5 network for segmenting medical image volumes with scribbles for training only. Our method is compared to five baselines and significantly outperforms scribble-based methods on three datasets. 
    \item We propose a label propagation module for 3D pseudo mask generation and an active boundary loss to regularize 3D segmentation results. These modules are general and could be used in other segmentation networks. 
\end{itemize}

\section{Scribble2D5: Scribbles-based Volumetric Segmentation}

Figure~\ref{fig:overview} presents the framework of our Scribble2D5, a weakly-supervised image segmentation network based on scribble annotations. Scribble2D5 has a 2.5D attention UNet~\cite{shapey2019artificial} as the backbone network, which is augmented by three modules, i.e., a label propagation module for generating 3D pseudo masks and boundaries, a static boundary prediction module  for incorporating object boundary information from images, and a segmentation boosting module for further considering active boundaries via an active boundary loss.

\subsection{3D Pseudo Label Generation via Label Propagation}
\label{sec:pseudolabel}

 Scribble annotations are often sparse, which cover a small amount of pixels on each slice of an image volume. As a result, the supervision information from scribbles is not strong enough to produce good guidance, like $\text{UNet}_\text{PCE}$~\cite{tang2018normalized}. To address this issue, we propose to magnify the effects of scribble annotations in 3D by leveraging supervoxels. That is, we adopt SLIC~\cite{achanta2012slic}, which generates supervoxels from images using an adaptive k-means clustering by considering both image intensity and distance similarities. We collect the supervoxels that scribbles pass through, resulting in 3D pseudo segmentation masks for ROIs.
 
 Except for the pseudo mask we generate from the scribble annotations, we generate the pseudo static boundary of ROI from an image volume by stacking 2D edges detected on each slice. This boundary is static since it is pre-computed from the image and keeps unchanged during training, which is different from the active boundary we will discuss later. To obtain 2D edges, we directly use an existing method, HED~\cite{xie2015holistically}, which is pretrained on the generic edges of BSDS500~\cite{arbelaez2010contour}. 
 In this way, we have a Label Propagation Module (LPM) to generate 3D pseudo labels from scribbles and images for ROI segmentation and pre-computed boundary for static boundary prediction, respectively. 

\subsection{Scribble2D5 Network}

\noindent
\textbf{Backbone.} The image volumes studied in the experiments have different voxel spacings. Roughly, the in-plane resolution within a slice is about four times the thickness of a slice. Since 2D CNNs ignore the important correlations among slices and 3D CNNs typically handle isotropic image volumes, we choose a 2.5D neural network that considers the anisotropic properties of an image volume. In particular, we adopt an attention UNet2D5~\cite{shapey2019artificial} as our backbone network, which augments UNet2D5 by adding a simple attention block at each deconvolutional layer, as shown in Fig.~\ref{fig:overview}. Specifically, at the top two layers of both encoder and decoder branches, we have 2D convolutional operations; while at other layers, the feature maps are isotropic, which are suitable for 3D convolutions. The attention blocks are colored in purple in Fig.~\ref{fig:overview}. Their attention maps are estimated via two layers of convolutions, i.e., one with ReLU and the other with a Sigmoid activation function. This 2.5D network suits for all images in our experiments. 

\noindent
\textbf{Static Boundary Prediction Module (SBPM).} This module encourages the backbone network to extract image features with rich boundary structures at different scales. Following~\cite{zhang2020weakly}, we collect feature maps from different layers of the network decoder, and concatenate them at different resolutions right after one convolutional layer with a filter of size $1 \times 1 \times 1$. To fuse these features, we feed them to a residual channel attention block (as shown by a green triangle in Fig.~\ref{fig:overview}) and a $1 \times 1 \times 1$ convolutional layer to produce a boundary map $b$ in 3D. Under the supervision of the previously generated 3D pseudo boundary $B$, the network is trained with a cross-entropy loss: 
$\mathcal{L}_{bry}(b, B)= -\sum_{c=1}^{N} B_c \textit{log}(b_c)$.
Here, $N$ is the total number of classes for segmenting in a dataset.

\noindent
\textbf{Segmentation Boosting Module (SBM).} This module performs segmentation under the supervision of the pseudo mask generated with supervoxels and a regularization on segmentation output. The module includes an initial segmentation and a final one with further considering static and active boundaries. To predict a preliminary mask, we employ a dense atrous spatial pyramid pooling (DenseASPP) block~\cite{yang2018denseaspp} right after the bottom layer of the backbone network in Fig.~\ref{fig:overview}, which enlarges its receptive fields by utilizing different dilation rates. In this block, the convolutional layers are connected in a dense way to cover a larger scale range without significantly increase the model size. 

To generate the initial segmentation mask $M^{init}$, we adopt two additional 3D convolutional layers followed by a $1 \times 1 \times 1$ convolution, resulting in the initial prediction supervised by the generated pseudo mask $M^{pseudo}$. Considering the oversegment nature of supervoxels, one supervoxel may be selected by multiple different classes. To avoid this confusion, we only consider those supervoxels with a unique label, which are set as 1 in the mask $M^{voxel}$ with others being zeros. We use the partial cross entropy to supervise the initial segmentation:  
$\mathcal{L}_{seg}(M^{init}, M^{pseudo}, M^{voxel}) = -\sum_{c=1}^{N} M^{voxel}_c \cdot M^{pseudo}_c \textit{log}(M^{init}_c)$.
This loss function allows early feedback to fasten the network convergence.

To refine the initial estimation and obtain a boundary-preserving mask for a final prediction, we merge SBPM outputs with those from the initial mask prediction for a refinement. These feature maps are fed to a residual channel attention block, followed by a $1 \times 1 \times 1$ convolutional layer to predict the final mask $M^{final}$. Similarly, we use the partial cross-entropy loss to predict the final mask under the supervision of the generated pseudo mask $M^{pseudo}$.

\noindent
\underline{Active Boundary (AB) Loss.} The pseudo masks are imperfect because supervoxels are coarse segmentation masks of ROIs and have oversegment issues, resulting in a potential of having many false positives. To mitigate this issue, we propose regularizing the surface and volume of the 3D segmentation region by upgrading the active contour loss~\cite{chen2019learning} to a 3D version. We apply an AB loss as following:
$\mathcal{L}_{AB} = \textit{Surface} + \lambda_{1} \cdot \textit{Volume}_\textit{In}  + \lambda_{2} \cdot \textit{Volume}_\textit{Out}$,
where
$\textit{Surface} =\int_{S}|\nabla u| \textit{ds}$ and $u$ is the prediction; $\textit{Volume}_\textit{In}=\int_{V}\left(c_{1}-v\right)^{2} \textit{udx}$, $c_1$ is the mean image intensity inside of interested regions $V$, and $v$ is the input image;  $\textit{Volume}_\textit{Out}=\int_{\bar{V}}\left(c_{2}-v\right)^{2} \textit{udx}$ and $c_2$ is the mean image intensity outside of the region. These items are balanced by two hyper-parameters $\lambda_1$ and $\lambda_2$. In the experiments, we set $\lambda_1=1$ and $\lambda_2=0.1$, to emphasis more on the inside region of the volume. This new loss function considers the shape and intensity of an image in 3D, which regularizes ROI's shapes and reduces false positives. 

\vspace{0.05in}
\noindent
The final loss function is 
    $\mathcal{L}_{total} = \beta_1 \mathcal{L}_{bry}(b, B) + \mathcal{L}_{seg}(M^{init}, M^{pseudo}, M^{voxel})  + \mathcal{L}_{seg}(M^{final}, M^{pseudo}, M^{voxel}) + \beta_2 \mathcal{L}_{AB}$.
Here, $\beta_1$ and $\beta_2$ are weights for balancing loss terms, which are both set as 0.3.

\newsavebox\CBox
\def\textBF#1{\sbox\CBox{#1}\resizebox{\wd\CBox}{\ht\CBox}{\textbf{#1}}}

\begin{table}[t]
\begin{center}
\scriptsize
\renewcommand\arraystretch{1.2} 

\caption{Quantitative comparison among baselines and our method for volumetric segmentation on three datasets. Mean and standard deviation (subscript) are reported. The upper bounds are colored in {\color{blue}blue}, and the best results by using scribbles are marked in \textbf{bold}. $^\dagger$P is short for Point, indicating extreme points. such annotations are available only in the VS dataset. $^*$These numbers are taken from~\cite{dorent2021inter}. (Best viewed in color)}

\begin{tabular}{p{0.2cm}|p{0.2cm}|p{2.1cm}|p{0.85cm}p{0.91cm}p{1.15cm}|p{0.85cm}p{0.91cm}p{1.15cm}|p{0.85cm}p{0.91cm}p{1.15cm}}
\toprule
\multicolumn{3}{c|}{\multirow{3}*{Approach}} &\multicolumn{3}{c|}{ACDC} & \multicolumn{3}{c|}{VS} & \multicolumn{3}{c}{CHAOS} \\ 
\cmidrule{4-12}
\multicolumn{3}{c|}{} & \makecell{Dice \\ (\%,$\uparrow$)} & \makecell{HD95 \\ (mm,$\downarrow$)} & \makecell{Precision \\(\%,$\uparrow$)} & \makecell{Dice \\ (\%,$\uparrow$)} & \makecell{HD95 \\ (mm,$\downarrow$)} & \makecell{Precision\\(\%,$\uparrow$)} & \makecell{Dice \\ (\%,$\uparrow$)} & \makecell{HD95 \\ (mm,$\downarrow$)} & \makecell{Precision \\(\%,$\uparrow$)} \\
\cmidrule{1-12}
\multirow{10}*{\begin{sideways} Supervision Type \end{sideways}} & \multirow{5}*{\begin{sideways} Scribble \end{sideways}} & $\rm UNet_{PCE}$~\cite{tang2018normalized} & \makecell[c]{$79.0_{06}$} & \makecell[c]{$6.9_{04}$} & \makecell[c]{$77.3_{06}$} & \makecell[c]{$44.6_{08}$} & \makecell[c]{$6.5_{03}$} & \makecell[c]{$43.8_{05}$} & \makecell[c]{$34.4_{06}$} & \makecell[c]{$9.4_{03}$} & \makecell[c]{$36.6_{05}$}  \\
& & MAAG~\cite{valvano2021learning}& \makecell[c]{$83.4_{04}$} & \makecell[c]{$8.6_{04}$} & \makecell[c]{$78.5_{05}$} & \makecell[c]{$69.4_{06}$} & \makecell[c]{$5.9_{05}$} & \makecell[c]{$56.8_{05}$} & \makecell[c]{$66.4_{05}$} & \makecell[c]{$3.8_{05}$} & \makecell[c]{$57.2_{06}$}  \\
\cmidrule{3-12}
& & Ours \tiny{w/o LPM}           & \makecell[c]{$83.2_{05}$} & \makecell[c]{$7.7_{03}$} & \makecell[c]{$84.1_{05}$} & \makecell[c]{$78.8_{05}$} & \makecell[c]{$4.6_{01}$} & \makecell[c]{$77.6_{05}$} & \makecell[c]{$81.2_{07}$} & \makecell[c]{$5.8_{08}$} & \makecell[c]{$82.0_{06}$}  \\
& & Ours \tiny{w/o SBPM}          & \makecell[c]{$85.6_{05}$} & \makecell[c]{$4.6_{04}$} & \makecell[c]{$85.5_{04}$} & \makecell[c]{$80.6_{05}$} & \makecell[c]{$7.1_{03}$} & \makecell[c]{{$\mathbf{81.6_{04}}$}} & \makecell[c]{$84.6_{05}$} & \makecell[c]{$5.5_{05}$} & \makecell[c]{$83.1_{05}$}  \\
& & Ours \tiny{w/o ABL}           & \makecell[c]{$88.7_{04}$} & \makecell[c]{$5.1_{08}$} & \makecell[c]{{$\mathbf{86.0_{05}}$}} & \makecell[c]{$81.0_{03}$} & \makecell[c]{$4.8_{01}$} & \makecell[c]{$80.1_{05}$} & \makecell[c]{$85.6_{04}$} & \makecell[c]{$4.8_{05}$} & \makecell[c]{$81.3_{02}$}  \\
& & Scribble2D5\tiny{(ours)}      & \makecell[c]{{$\mathbf{90.6_{03}}$}} & \makecell[c]{{$\mathbf{2.3_{05}}$}} & \makecell[c]{$84.7_{05}$} & \makecell[c]{{$\mathbf{82.6_{07}}$}} & \makecell[c]{{${4.7_{04}}$}} & \makecell[c]{$81.5_{06}$} & \makecell[c]{{$\mathbf{86.0_{04}}$}} & \makecell[c]{{$\mathbf{2.9_{02}}$}} & \makecell[c]{{$\mathbf{88.2_{03}}$}}  \\
\cmidrule{2-12}
& \begin{sideways}$\text{P}^{\dagger}$\end{sideways} & InExtremeIS~\cite{dorent2021inter}   & \makecell[c]{-} & \makecell[c]{-} & \makecell[c]{-} & \makecell[c]{$81.9_{03}^{*}$} & \makecell[c]{\color{blue}{{$\mathbf{3.7_{03}}^{*}$}}} & \makecell[c]{\color{blue}{$92.9_{02}^{*}$}}  & \makecell[c]{-} & \makecell[c]{-} & \makecell[c]{-} \\ 
\cmidrule{2-12}
& \multirow{2}*{\begin{sideways} Mask \end{sideways}} & 2D UNet~\cite{ronneberger2015u} & \makecell[c]{$93.0_{05}$} & \makecell[c]{$3.5_{15}$} & \makecell[c]{$90.2_{07}$} & \makecell[c]{$80.4_{03}$} & \makecell[c]{$7.3_{04}$} & \makecell[c]{$81.2_{03}$} & \makecell[c]{$82.3_{04}$} & \makecell[c]{$3.3_{01}$} & \makecell[c]{$81.7_{05}$} \\
& & 2.5D UNet~\cite{shapey2019artificial} & \makecell[c]{\color{blue}{$96.1_{03}$}} & \makecell[c]{\color{blue}{$0.3_{00}$}} & \makecell[c]{\color{blue}{$95.3_{04}$}} & \makecell[c]{\color{blue}{$87.3_{02}$}} & \makecell[c]{$6.8_{04}$} & \makecell[c]{$84.7_{03}$} & \makecell[c]{\color{blue}{$90.8_{03}$}} & \makecell[c]{\color{blue}{$1.1_{00}$}} & \makecell[c]{\color{blue}{$91.4_{05}$}} \\
\bottomrule
\end{tabular}

\vspace{-0.1in}
\label{tab:results}
\end{center}
\end{table}

\begin{figure}[t]
\includegraphics[width=\textwidth]{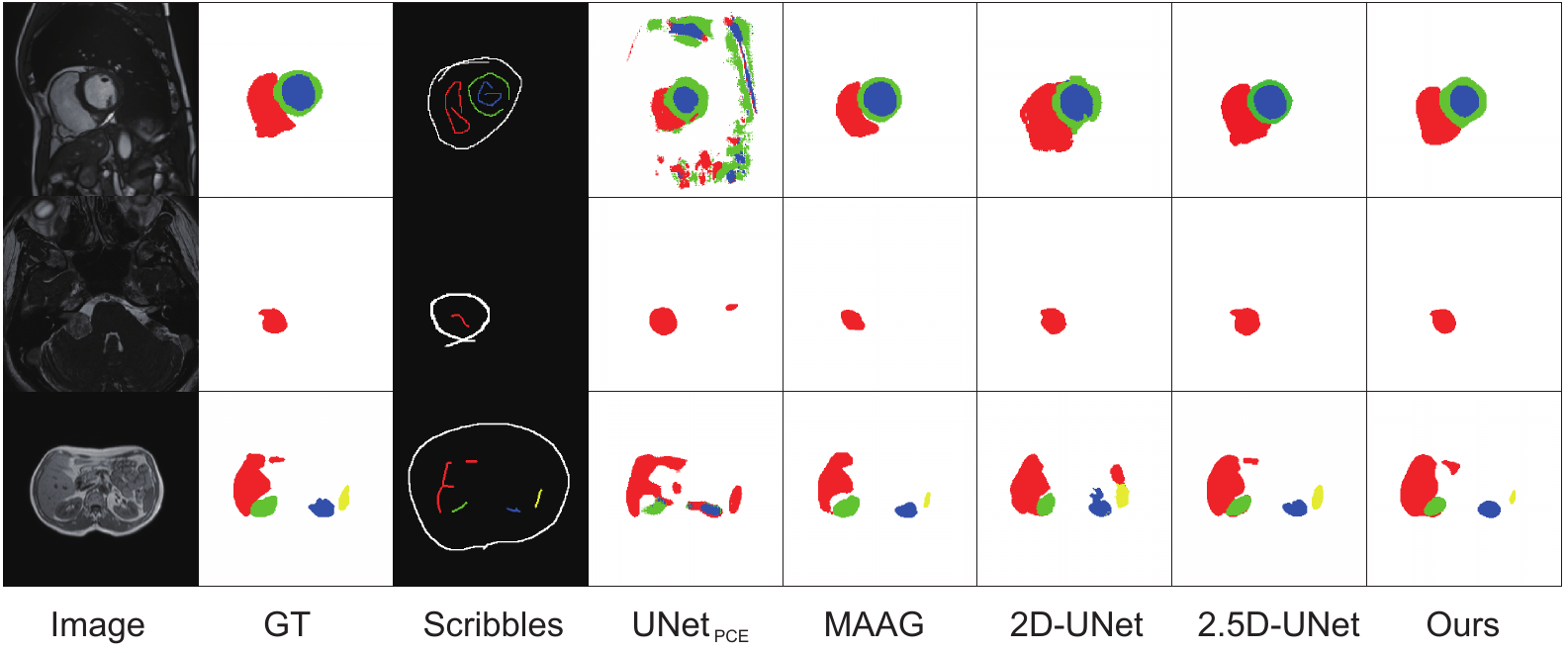}
\caption{Qualitative comparison among scribble-based ($\text{UNet}_\text{PCE}$ and MAAG), mask-based (2D and 2.5D UNets), and our methods. (Best viewed in color)} 
\label{fig:qualitive}
\end{figure}

\section{Experiments}
\subsection{Datasets and Experimental Settings}
\noindent 
\textbf{ACDC Dataset~\cite{bernard2018deep}.}
This dataset consists of Cine MR images collected from 100 patients. 
Manual segmentation masks of the left and right ventricles and myocardium are provided at the end-diastolic and end-systolic cardiac phases. The slice size is $256 \times 208$ with the pixel spacing varying from 1.37 to 1.68$mm$. The number of slices is between 28 and 40, and the slice thickness is 5$mm$ or 8$mm$. We subject-wisely divide the ACDC dataset into sets of 70\%, 15\% and 15\% for training, validation, and test, respectively.


\noindent
\textbf{VS Dataset~\cite{shapey2021segmentation}.}
This dataset collects T2-weighted MRIs from 242 patients with a single sporadic vestibular schwannoma (VS) tumor.
The size of an image slice is $384 \times 384$ or $448 \times 448$, with a pixel spacing of $~0.5 \times 0.5mm^2$. The number of slices varies from 19 to 118, with a thickness of $1.5mm$. The VS tumor masks are manually annotated by neurosurgeons and physicists. The dataset is subject-wisely split into 172 for training, 20 for validation, and 46 for test. 

\noindent
\textbf{CHAOS Dataset~\cite{kavur2021chaos}.}
This dataset has abdominal T1-weighted MR images collected from 20 subjects and the corresponding segmentation masks for liver, kidneys, and spleen. The image slice size is $256 \times 256$ with a resolution of  $1.36-1.89mm$ (average $1.61mm$). The number of slices is between 26 and 50 (average 36) with the slice thickness varying from 5.5 to 9$mm$ (average 7.84 $mm$). We also subject-wisely divide this dataset into sets of 70\%, 15\% and 15\% for training, validation, and test, respectively.

\noindent
\textbf{Scribble Generation and Other Settings.} For the ACDC dataset, we use the scribbles provided in~\cite{valvano2021learning}, whch are manually drawn by experts at both end-diastolic and end-systolic phases. For both VS and CHAOS datasets, following~\cite{rajchl2017employing}, we simulate scribbles by an iterative morphological erosion and closing of segmentation masks, which results in a one-pixel skeleton for each object. Since the resulting background scribble is winding, we use ITK-Snap to annotate background with 1-pixel width curves.

For all datasets, we randomly crop an image volume and obtain patches of size $224 \times 224 \times 32$ as the network inputs. An image volume is padded with zeros if its size is smaller than the input size. 

We train all models for 200 epochs with early stopping. 
The weights of the network are initialized by following a normal distribution with a mean of 0 and variance of 0.01. We use Adam optimizer with a weight decay $10^{-7}$ and an initial learning rate 1$e$-4. The whole training takes about 6 hours with a batch size of 4 on one NVIDIA GeForce RTX 3090 GPU.

\noindent
\textbf{Baselines and Evaluation Metrics.}
To demonstrate the effectiveness of our methods, we select three groups of baselines, including fully-supervised methods (i.e., 2D UNet~\cite{ronneberger2015u} and 2.5D UNet~\cite{shapey2019artificial}), weakly-supervised methods using scribbles (i.e., $\text{UNet}_\text{PCE}$~\cite{tang2018normalized} and MAAG~\cite{valvano2021learning}) and a weakly-supervised method using extreme points~\cite{dorent2021inter}. To evaluate the segmentation performance, we use the Dice score to calculate the overlap between our segmentation and the ground truth (GT), the 95th percentile of the Hausdorff Distance (HD95) to measure the distance between our ROI boundary and GT, and the precision to check the purity of the positively-segmented voxels. 

\subsection{Experimental Results}
Table~\ref{tab:results} presents our experimental results on three datasets with comparison to five baselines. For all datasets, the upper bounds of the segmentation performance are mainly provided by the 2.5D UNet, which are colored in blue in Table~\ref{tab:results}. Compared to the scribble-based SOTA method on ACDC and CHAOS datasets, i.e., MAAG~\cite{valvano2021learning}, scribble2D5 improves the Dice score by $7\%$ and $19.5\%$, reduces the HD95 by $6.3$mm and $1.8$mm, and improves the precision by $6.2\%$ and $29.5\%$, respectively. In addition, our method outperforms the most recent two methods, i.e., ScribbleSeg~\cite{luo2022scribble} and CycleMix~\cite{zhang2022cyclemix}, on ACDC Dataset, with our 0.903 mean dice vs 0.872 in~\cite{luo2022scribble} (using the same 5-fold cross validation), our 0.896 mean dice vs 0.848 in~\cite{zhang2022cyclemix} (similarly, using 35 subjects for training). Compared to the extreme-point-based SOTA method on the VS dataset, i.e., InExtremeIS~\cite{dorent2021inter}, although our method has a lower precision and HD95 value, it improves the Dice score by $0.7\%$.
We do not report InExtremeIS' results on ACDC and CHAOS datasets because extreme points for these two datasets are not available or easy to be generated. 

Figure~\ref{fig:qualitive} demonstrates the sampled qualitative results of our method compared to the baselines. Overall, we have fewer false positives compared to scribble-based methods, i.e., $\text{UNet}_\text{PCE}$ and MAAG, and better boundary localization with more accurate boundary prediction for each ROI. Regarding the comparison with mask-based methods, our method sometimes generates even better masks than 2D UNet, while needs improvements at details compared to 2.5D UNet.

\begin{figure}[t]
\includegraphics[width=\textwidth]{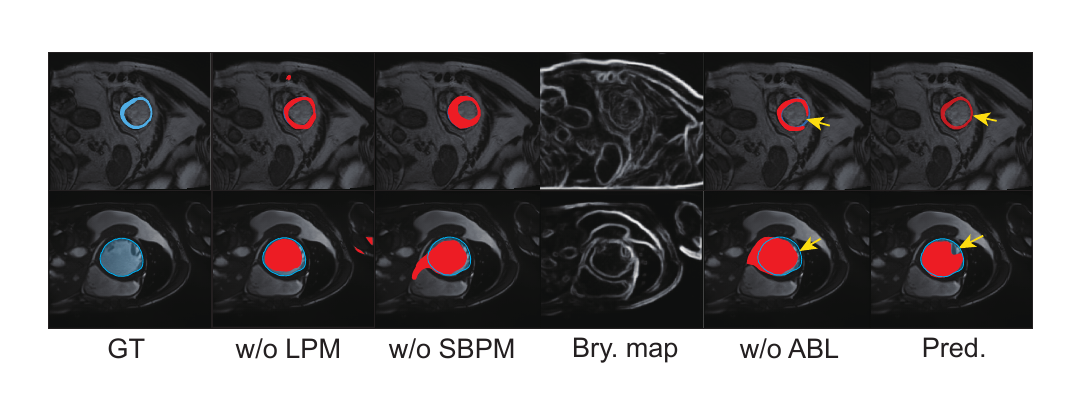}
\caption{Visualization of our intermediate and final results on image sampels from ACDC dataset. The ground truth (GT) is colored in blue, like the blue region in the first column and the blue contours in other images, while our predictions are colored in red. Yellow arrows show the effect of the active boundary loss (ABL). (Best viewed in color)}
\label{fig:ablation}
\end{figure}

\noindent
\textbf{Ablation Study.}
To check the effectiveness of each module in our method, we perform an ablation study with three variants: 
a) \textbf{Ours w/o LPM}: Scribble2D5 without the label propagation module (LPM);
b) \textbf{Ours w/o SBPM}: Scribble2D5 without the static boundary prediction model (SBPM), including SBPM and active boundary loss; and
c) \textbf{Ours w/o ABL}: Scribble2D5 without the active boundary loss (ABL). Take the ACDC dataset as an example, as shown in Table~\ref{tab:results}, without LPM but all others, the Dice score reduces from $90.4\%$ to $80.6\%$. With LPM but without SBPM, the Dice score is $85.6\%$; then the static boundary prediction module contributes an improvement of $3.1\%$, and the active boundary loss contributes an additional improvement of $1.7\%$ in Dice score. Figure~\ref{fig:ablation} visualizes two samples from the ACDC dataset with our intermediate and final prediction results. Without LPM, our method suffers from false positives far away from the ROI; without SBPM, our method has oversegment issues of the ROI; by adding the boundary map and active boundary regularization, our method adjusts the prediction based on the image edge and texture information, resulting the closest results compared to GT.    

\section{Conclusion and Discussion}
In this paper, we proposed a weakly-supervised volumetric image segmentation network, Scribble2D5, which significantly outperforms existing scribble-based methods and reduces the performance gap between weakly-supervised and full-supervised segmentation methods. One limitation of our method is that our pseudo boundary labels are not purely 3D, which will be explored in the future. Currently, we do not consider the case of missing scribble annotations on some slices or further reducing the manual work via an adaptive annotation. One potential solution for this is using the watershed techinque~\cite{vincent1991watersheds} in 3D. Also, We observe that the shape and location of scribbles would affect the segmentation accuracy, summarizing a way to make scribble annotations for different shapes of ROIs will be useful in practice by providing some rules for users to make annotations, which will benefit the segmentation and be left as the future work.

\subsubsection{Acknowledgment.}
This work was supported by Shanghai Municipal Science and Technology Major Project 2021SHZDZX0102.

%
%
%
\bibliographystyle{splncs04}
\bibliography{refs.bib}

\end{document}